%% file: paper.tex
\documentclass[final,journal,twocolumn]{IEEEtran}

\usepackage{graphicx}
\usepackage[tight]{subfigure}
\usepackage[cmex10]{amsmath}
\usepackage{amssymb,amsthm}
\usepackage{array}
\usepackage{booktabs} 
\usepackage{adjustbox}
\usepackage{xcolor}
\usepackage{cite}
\usepackage{pifont}
\usepackage[normalem]{ulem}
\usepackage{multirow}
\usepackage{tikz}
\usepackage[hidelinks=true]{hyperref}

\graphicspath{{figures/}}
\DeclareGraphicsExtensions{.pdf,.jpg,.png}

\input{macros}

\begin{document}

\title{Multisource Region Attention Network\\
    for Fine-Grained Object Recognition\\ in Remote Sensing Imagery%
    \thanks{This work was supported in part by the TUBITAK Grant 116E445,
    BAGEP Award of the Science Academy, and METU Research Fund Project 2744.}%
}
\author{%
    Gencer~Sumbul,~\IEEEmembership{Student Member,~IEEE},
    Ramazan~Gokberk~Cinbis,
    and~Selim~Aksoy,~\IEEEmembership{Senior Member,~IEEE}%
    \thanks{G. Sumbul and S. Aksoy are with the Department of Computer Engineering,
    Bilkent University, Ankara, 06800, Turkey. After this work has been done,
    the affiliation of Gencer Sumbul has changed to TU Berlin.
    Email: \mbox{gencer.suembuel@tu-berlin.de}, \mbox{saksoy@cs.bilkent.edu.tr}.}%
    \thanks{R. G. Cinbis is with the Department of Computer Engineering,
    METU, Ankara, 06800, Turkey.
    Email: \mbox{gcinbis@ceng.metu.edu.tr}.} %
    \thanks{\textsuperscript{\textcopyright} 2019 IEEE. Personal use of this material is permitted. Permission from IEEE
    must be obtained for all other uses, in any current or future media, including 
    reprinting/republishing this material for advertising or promotional purposes, creating new 
    collective works, for resale or redistribution to servers or lists, or reuse of any copyrighted 
    component of this work in other works.} %
}

\markboth{TO APPEAR IN IEEE Transactions on Geoscience and Remote Sensing, 2019}{Sumbul \MakeLowercase{\textit{et al.}}: Multisource Region Attention Network}

\maketitle

\input{abstract}

\input{intro}
\input{dataset}
\input{method}

\input{experiments}
\input{conclusions}

\bibliographystyle{IEEEtran}
\bibliography{definitions,bibliography}

\input{bio}

\end{document}

%% file: macros.tex
\usepackage{color}
\usepackage[normalem]{ulem}

\usepackage{amsmath,amssymb,amsbsy,xspace}
\makeatletter
\DeclareRobustCommand\onedot{\futurelet\@let@token\@onedot}
\def\@onedot{\ifx\@let@token.\else.\null\fi\xspace}

\def\ie{{i.e}\onedot, } 
 
\def\etc{{etc}\onedot}

\def\be{\begin{equation}}
\def\ee{\end{equation}}
\def\bea{\begin{eqnarray}}
\def\eea{\end{eqnarray}}

\def\sect#1{Section~\ref{sec:#1}}

\makeatother

%% file: abstract.tex
\begin{abstract}
Fine-grained object recognition concerns the identification of the type of an
object among a large number of closely related sub-categories. Multisource data
analysis, that aims to leverage the complementary spectral, spatial, and
structural information embedded in different sources, is a promising direction
towards solving the fine-grained recognition problem that involves low
between-class variance, small training set sizes for rare classes, and class imbalance.
However, the common assumption of co-registered sources may not hold at the
pixel level for small objects of interest. We present a novel methodology that
aims to simultaneously learn the alignment of multisource data and the
classification model in a unified framework. The proposed method involves a
multisource region attention network that computes per-source feature
representations, assigns attention scores to candidate regions sampled around
the expected object locations by using these representations, and classifies
the objects by using an attention-driven multisource representation that
combines the feature representations and the attention scores from all sources.
All components of the model are realized using deep neural networks and are
learned in an end-to-end fashion. Experiments using RGB, multispectral, and
LiDAR elevation data for classification of street trees showed that our approach
achieved $64.2\%$ and $47.3\%$ accuracies for the $18$-class and $40$-class
settings, respectively, which correspond to $13\%$ and $14.3\%$ improvement
relative to the commonly used feature concatenation approach from multiple
sources.
\end{abstract}


%% file: intro.tex
\section{Introduction}
\label{sec:Introduction}

New generation sensors used for remote sensing has allowed the acquisition of
images at very high spatial resolution with rich spectral information.
A challenging problem that has been enabled by such advances in sensor
technology is \emph{fine-grained object recognition} that involves the identification
of the type of an object in the domain of a large number of closely related
sub-categories. This problem differs from the traditional object recognition
and classification tasks predominantly studied in the remote sensing literature
in at least three main ways: (i) differentiating among many similar categories
can be much more difficult due to low between-class variance, (ii) difficulty
of accumulating examples for a large number of similar categories can greatly
limit the training set sizes for some classes, (iii) class imbalance can
cause the conventional supervised learning formulations to overfit to more
frequent classes and ignore the ones with limited number of samples. Such
major differences lead to an uncertainty in the applicability of existing
approaches developed based on traditional settings. Thus, the development
of methods and benchmark data sets for fine-grained classification
is an open research problem, whose importance is likely to increase over time.

Fine-grained object recognition has received very little attention in the
remote sensing literature. Oliveau and Sahbi \cite{Oliveau2017} proposed an
alternating optimization procedure that iteratively learned a dictionary-based
attribute representation and a support vector machine (SVM) classifier based on
these attributes for classification of image patches into 12 ship categories.
Branson et al.\ \cite{Branson:2018} jointly used
aerial images and street-view panoramas for fine-grained classification of
street trees. They concatenated the feature representations computed by deep
networks independently trained for the aerial and ground views, and fed these
features to a linear SVM for classification of 40 tree species.
In \cite{Sumbul:2018}, we studied the more extreme zero-shot learning scenario
where no training example exists for some of the classes. First, a compatibility
function between image features extracted from a convolutional neural network
(CNN) and auxiliary information about the semantics of the classes of interest
was learned by using samples from the seen classes. Then, recognition was done
by maximizing this function for the unseen classes. Experiments were done by
using an RGB image data set of 40 street tree categories.

New approaches that aim to learn classifiers under the presence of low
between-class variance, small sample sizes for rare classes, and class imbalance can
overcome these problems by enriching the data sets so that increased
spectral and spatial content provides potentially more identifying information that can be
exploited for discriminating instances of fine-grained classes. However, these two types of
information do not necessarily come together in the same data source. Thus,
multisource remote sensing is a promising research direction for fine-grained
recognition. For example, very high spatial resolution RGB data provides
texture information, whereas multi- and hyperspectral images
contain richer spectral content. Furthermore, LiDAR-based elevation models can
provide complementary information about the heights and other structural
characteristics of the objects.

Multisource image analysis \cite{Chova:2015} has been a popular problem in
remote sensing with a wide range of solutions including dependence trees
\cite{Datcu:2002}, kernel-based methods \cite{Valls:2008}, copula-based
multivariate statistical model \cite{Voisin:2014}, active learning
\cite{Zhang:2015}, and manifold alignment \cite{Tuia:2014,Gao:2018}.
Combining information from multiple data sources has also been the focus of
data fusion contests \cite{Debes2014,Liao:2015,Taberner:2016,Yokoya:2018} for land
cover/use classification. Similar to their popularity in general classification
tasks, deep learning-based approaches also received interest in multisource
analysis. Deep networks have typically been used in the classification stage
where raw optical bands and LiDAR-based digital surface models (DSM)
\cite{Morchhale2016,Pibre:2017} as well as handcrafted features from hyperspectral
and LiDAR data \cite{Ghamisi:2017} were concatenated and given as input to a
CNN classifier, or in the feature extraction stage
where independently learned deep feature representations from hyperspectral
and SAR data \cite{Hu:2017} or hyperspectral and LiDAR data \cite{Xu:2018}
were concatenated to form the input of a separate classifier. The output of a
fully convolutional network trained on optical data and a logistic regression
classifier trained on LiDAR data were also used in decision-level fusion
by using a conditional random field \cite{Liu:2017}.

A common assumption in all of these approaches is that the data sources are
georeferenced or co-registered so that concatenation can be used for pixel-wise
classification. Potential registration errors may not cause a problem during
learning if one considers a small number of relatively distinct classes with many
samples, or during testing where evaluation is done by pixels sampled from the
inside of large regions labeled by land cover/use classes. Even though various
approaches have been proposed for the registration of multisensor images
\cite{Lee:2015,Marcos:2016}, finding pixel-level correspondences between images
acquired from different sensors with different spatial and spectral resolution
may not be error-free due to differences in the imaging conditions, viewing
geometry, topographic
effects, and geometric distortions \cite{Han:2016}. Furthermore, errors at the
level of a few pixels may not matter when the goal is to classify pixels
sampled from land cover/use classes such as road, building, vegetation, soil,
water, but can be very significant for fine-grained object recognition when the
objects of interest, e.g., individual trees as in this paper, can appear
as small as a few pixels even in very high spatial resolution images.

In this paper, we propose a multisource fine-grained object recognition methodology
that aims to simultaneously learn the \emph{alignment} of the images acquired
from different sources and the \emph{classification} model in a unified framework.
We illustrate this framework in the fine-grained categorization of 40 different
types of street trees using data from RGB, multispectral (MS) and LiDAR sensors.
Classification of urban tree species provides a suitable test bed for this
challenging fine-grained recognition problem because of the
difficulty of finding field examples for training data, fine-scale spatial
variation, and high species diversity \cite{Liu2017}. Furthermore,
appearance variations with respect to scale and spectral values, the difficulty of
co-registration of small objects of interest in multiple sources, and rareness
of certain species resemble the typical problems in fine-grained object
recognition \cite{Fassnacht2016}. Consequently, differentiating the
sub-categories can be a very difficult task even with visual inspection
using very high spatial resolution imagery. Classification of tree species
has been previously studied in the remote sensing literature by using
specialized approaches via fusion of hyperspectral and LiDAR data with linear
discriminant \cite{Alonzo2014}, nearest neighbor \cite{Voss2008}, SVM
\cite{Dalponte:2012}, or random forest \cite{Naidoo2012,Dalponte:2012,Liu2017}
classifiers. However, all of these approaches were specialized to tree
classification, and none of them considered the alignment problem.

\begin{figure}
\centering
\includegraphics[width=0.88\linewidth]{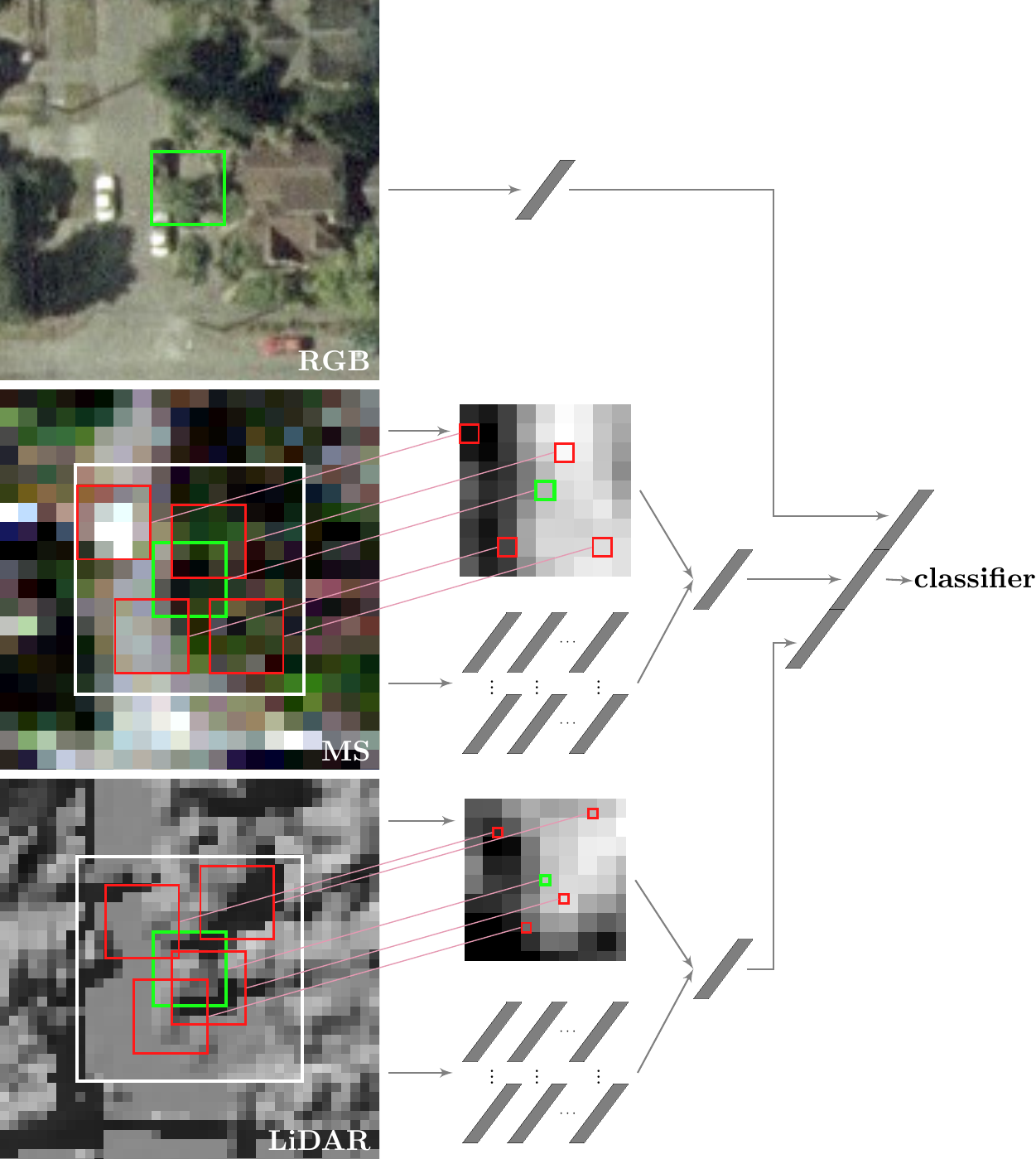}
\caption{Attention-driven representation of multisource data with imperfect
pixel-level alignment. The RGB image, where a $25 \times 25$ pixel region (shown
as green) centered at a verified ground truth tree location, is considered as
the \emph{reference} source. The corresponding regions 
in the MS and LiDAR data occupy $4 \times 4$ and $8 \times 8$ pixels
(also shown as green), respectively. The proposed \emph{multisource region
attention network} evaluates candidate regions (shown as red) sampled within
a larger neighborhood (shown as white) around the expected object position,
computes \emph{attention scores} (shown as a grayscale map) that represent the
confidence about each candidate containing the object, uses these scores to
form the \emph{attention-driven multisource representation}.}
\label{fig:Alignment}
\end{figure}

Our main contributions in this paper are as follows. The first contribution is
a novel \emph{multisource region attention network} that simultaneously learns
to \emph{attend} regions of source images that are likely to contain the object
of interest and to \emph{classify} the objects by using an \emph{attention-driven
multisource deep feature representation}. The problem flow is illustrated in
Figure \ref{fig:Alignment}. We assume that the objects exist in all sources but
their exact positions are unknown except one particular source, the
\emph{reference}, that is verified with respect to the ground truth. The
proposed network learns to (i) compute per-source deep feature representations,
(ii) assign attention scores to candidate regions sampled around the expected
object locations in remaining sources by relating them to the reference, and
(iii) classify objects by using an attention-driven multisource representation
that combines the feature representations and the attention scores from all
sources. A deep neural network architecture is presented to realize the
components of this framework, and learn them in an end-to-end fashion. The
second contribution is the detailed evaluation of this framework by using
different combinations of source images from RGB, MS, and LiDAR
sensors in fine-grained categorization of $40$ different types of tree species.
To the best of our knowledge, the proposed methodology is the
first example for a generic unified framework for fine-grained object
recognition by using any number of sources with different spatial and spectral
resolutions via simultaneous learning of alignment and classification models.

The rest of the paper is organized as follows. Section \ref{sec:DataSet}
introduces the fine-grained data set. Section \ref{sec:Methodology} describes
the details of the methodology. Section \ref{sec:Experiments} presents the
experiments. Section \ref{sec:Conclusions} provides the conclusions.

%% file: dataset.tex
\section{Data set}
\label{sec:DataSet}

The data set in \cite{Sumbul:2018} contained $48,\!063$ instances of street
trees belonging to $40$ categories. Each instance was represented by
an aerial RGB image patch of $25 \times 25$ pixels at $1$ foot spatial
resolution, centered at points provided in the point GIS data. The names of the
classes and the number of samples can be found in \cite{Sumbul:2018}. We use both
an $18$-class subset (named the supervised set in \cite{Sumbul:2018}) and
the full set of $40$ classes here.

This work extends that data set with an $8$-band \mbox{WorldView-2} MS image and
a LiDAR-based DSM with $2$ meter and $3$ foot spatial resolution, respectively.
Consequently, each tree instance corresponds to a $4 \times 4$
pixel patch in the MS data, and an $8 \times 8$ pixel patch in the LiDAR data.
Since the RGB image has the highest spatial resolution and the corresponding
annotations were verified by visual inspection in \cite{Sumbul:2018}, we
consider it as the \emph{reference} source. Even though each source image was
previously georeferenced, precise pixel-level alignments among these sources
were not possible as shown in Figure \ref{fig:Alignment}. Thus, the proposed
methodology in the following section aims to find the true, yet unknown,
matching patch of $4 \times 4$ pixels in the neighboring region of $12 \times 12$
pixels in the MS data, and the corresponding patch of $8 \times 8$ pixels
within a $24 \times 24$ pixel region in the LiDAR data.
The neighborhood sizes are selected empirically using validation data.
Using larger neighborhoods risks the inclusion of other trees
that can confuse the attention mechanism, and smaller
neighborhoods may not contain sufficient number of candidates.

%% file: method.tex
\section{Methodology}
\label{sec:Methodology}

In this section, we first introduce the multisource object recognition problem and present a baseline scheme for it.
Then, we explain our Multisource Region Attention Network approach, followed by the details of
the network architecture.

\subsection{Multisource object recognition problem}

In the multisource object recognition problem, we assume that there exists $M$
different source domains, where the space of samples from the $m$-th domain is
represented by $\mathcal{X}^m$.  Our goal is to learn a classification function
that maps a given object represented by a tuple of input instances from the
source domains $(x^1 \in \mathcal{X}^1,...,x^{M} \in \mathcal{X}^M)$ to one of
the classes $y \in \mathcal{Y}$ where $\mathcal{Y}$ is the set of all classes.

In this work, we focus on the problem of object recognition from multiple
source images, where each source corresponds to a particular sensor, such
as RGB, MS, LiDAR, \etc. We are particularly interested in the utilization of
overhead imagery, where the samples are typically collected from cameras with
different viewpoints, elevations, resolutions, dates and time of day. Such
differences in imaging conditions across the data sources make the precise
spatial alignment of the images very difficult. 
The image contents may also differ due to changes in
the area over time and occlusions in the scene.

In the next section, we first present a simple baseline approach towards
utilizing such multiple sources, and then, we explain our approach for
addressing these challenges in a much more rigorous way.

\subsection{Multisource feature concatenation}
\label{sec:basic_model}

\begin{figure*}
\begin{minipage}[b]{0.29\linewidth}
\subfigure[]{%
    \label{fig:basic_model}
    \includegraphics[width=\linewidth]{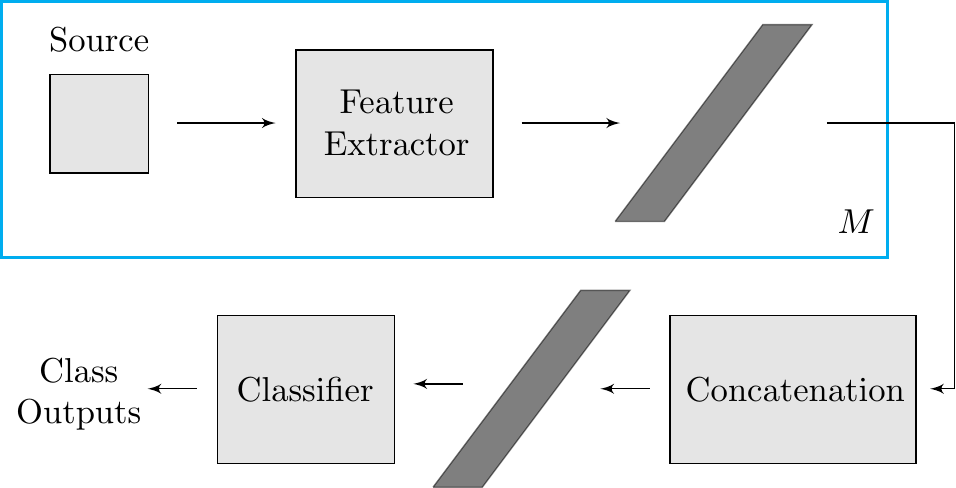}}
\end{minipage}
\hfill
\begin{minipage}[b]{0.66\linewidth}
\subfigure[]{%
    \label{fig:weight_model}
    \includegraphics[width=\linewidth]{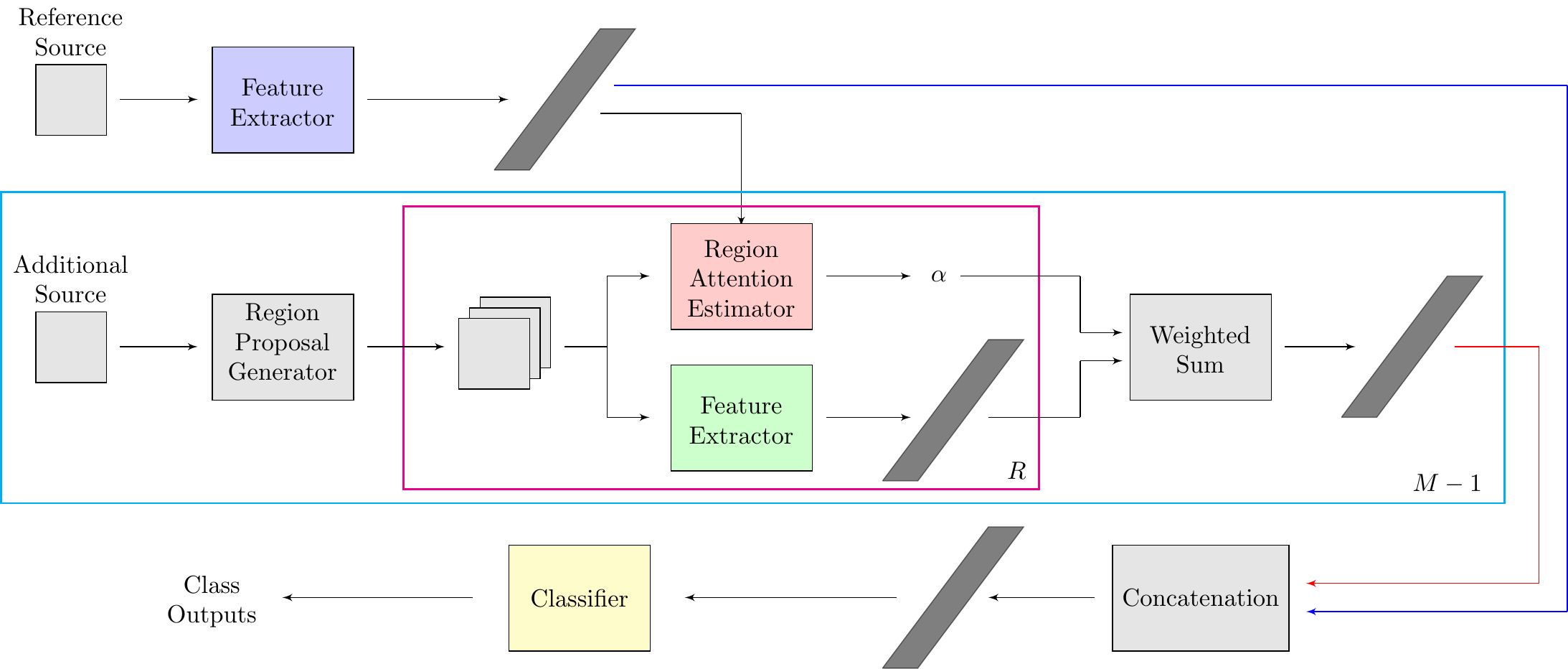}}
\end{minipage}
\caption{(a) Basic multisource model. The feature representations independently
obtained from each source are concatenated as the object representation.
(b) The proposed multisource region attention network. One source is chosen as
the reference with verified selection of its object locations according to
the ground truth. The remaining sources with the possibility of imprecise
alignment of object locations are considered as additional sources. For each
such source $x^m$, region proposals (candidate patches) $\{ x_1^m, \ldots,
x_R^m \}$ are generated, feature representations $\phi^\text{reg}_m(x_r^m)$
are obtained, and the attention scores (weights) $\alpha_m( x_r^m )$ are
computed with the help of the feature representation of the reference source.
The final representation for each additional source is the weighted sum
$\phi^\text{att}_m(x^m)$ of its proposal regions' representations, and the
final representation $\phi^\text{att}(x)$ used for class prediction is
obtained by concatenation.
Even though alternative weighting schemes such as filter weighting
\cite{Lu:2017} are possible, we focus on the weighting of the region proposals
in this paper.}
\label{fig:multisource_models}
\end{figure*}

A simple and commonly used scheme for utilizing multiple images in classification is to
extract features independently across the images and then concatenating them later, which is often called
{\em early fusion}. More precisely, for each source $m$, we assume that there exists a feature extractor
$\phi_m(x^m)$ which maps the input $x^m$ to a $d_m$-dimensional feature
vector. 
In this approach, it is presumed that each multisource tuple $x=(x^1,...,x^M)$ consists of the images of the same object from all sources,
and these images are spatially registered. Then, the multisource representation $\phi(x)$
is obtained by concatenating per-source feature vectors:
\begin{equation}
    \phi(x) = [\phi_1(x^1)^\top,...,\phi_{M}(x^{M})^\top]^\top.
    \label{eq:fused_im_embed}
\end{equation}
Once the multisource representation is obtained, the final object class
prediction is given by a classification function. This approach is illustrated
using plate notation\footnote{The plate notation represents the variables that
repeat in the model where the number of repetitions is given by the number on
the bottom-right corner of the corresponding rectangle enclosing these variables.}
in Figure~\ref{fig:basic_model}.

The main assumption of the simple feature concatenation approach is that the representation
obtained independently from each source successfully captures the characteristics of the object
within the target region. However, registration across the sources is usually imprecise,
which requires choosing relatively large regions to ensure that all images within a tuple contain the same object
instance.  In this case, however, the features extracted from these relatively large regions are likely to be dominated
by background information, which can greatly degrade the accuracy of the final classification model.

This problem is tackled by the proposed Multisource Region Attention Network, explained in the following section.

\subsection{Multisource Region Attention Network (MRAN)}
\label{sec:MRAN}

A central problem in multisource remote sensing is the difficulty of registration of source images, as discussed above.
To address this problem, we propose a deep neural network that learns to {\em attend} regions of source images such
that the resulting multisource representation is most informative for recognition purposes. We refer to this approach
as {\em Multisource Region Attention Network} (MRAN).

In our approach, we presume that there is (at least) one source for which feature extraction is reliable, \ie
the representation for this source is not dominated by noise and/or background information. 
We refer to this source as the {\em reference}.
Since the reference source typically has a higher spatial resolution,
it is often possible to annotate objects in it with high spatial fidelity,
either by geo-registering it with some form of ground truth or by visually
inspecting the images.

Our goal is to enhance recognition by leveraging additional sources. While we presume that all images within a tuple
contain the same object instance, we do not expect a precise spatial alignment among them,
\ie the exact position of an object is locally unknown in the sources other than the reference.
In addition, the images of additional sources
may potentially contain other object instances belonging to different classes.
In this realistic setting, therefore, extracting features independently at each source
is likely to perform poorly.

In our approach, we aim to overcome these difficulties via a deep network including a conditional attention mechanism
that selectively assigns importance scores to regions in each one of the $M-1$ sources, \ie in those other than the
reference.  Without loss of generality, we assume that the reference source is the very first one, and there are
$R$ candidate regions in each one of the other source images, denoted by $x^m_1,...,x^m_R$ where $m\in\{2,...,M\}$.
In our experiments, we obtain these candidate regions (proposals) by regularly sampling overlapping patches of fixed size
within a larger neighborhood around the expected position of the object obtained
by a simple transformation from the reference source
(see Section \ref{sec:Experiments} for details).

To formalize the proposed conditional attention mechanism, we define the
\emph{region attention estimator} $\omega_m(x^m_r,\phi_1(x^1))$, which takes
the $r$-th candidate region from the $m$-th source and the feature
representation of the corresponding reference image,
and maps to a non-negative attention score. The attention score represents the confidence that the region
contains (a part of) the object of interest. 

We then leverage these scores to obtain a multisource representation
that focuses on the regions containing the object of interest within each source. For this purpose, we define the
{\em attention-driven source representation}
$\phi^\text{att}_m(x^m), m = 2, \ldots, M$, as a weighted sum of per-region representations:
\begin{equation}
    \phi^\text{att}_m(x^m) = \sum_{r=1}^R\alpha_m(x_r^m)\phi^\text{reg}_m(x_r^m) ,
    \label{eq:est_im_embed}
\end{equation}
where $\phi^\text{reg}_m$ is the region-level feature extractor, and
the weighting term $\alpha_m$ is the normalized attention score of the region:
\begin{equation}
    \alpha_m(x_r^m) = \frac{\omega_m(x^m_r,\phi_1(x^1))}{\sum_{r'=1}^R \omega_m(x^m_{r'},\phi_1(x^1))}.
\end{equation}
Our final \emph{attention-driven multisource representation} is obtained by concatenation of attention-driven source
representations:
\begin{equation}
    \phi^\text{att}(x) = [\phi_{1}(x^{1})^\top, \phi^\text{att}_2(x^2)^\top,...,\phi^\text{att}_M(x^{M})^\top]^\top .
    \label{eq:new_fused_im_embed}
\end{equation}
The resulting attention-driven multisource representation can be fed to a
classifier to recognize the object of interest:
\begin{equation}
    C( x ) : \phi^\text{att}(x) \rightarrow \mathbb{R}_{\geq 0}^K
\end{equation}
where $C$ is the classifier function that outputs a confidence score for each
of the $K$ classes. An illustration of our MRAN framework can be found in
Figure \ref{fig:weight_model}.

In the next section, we present the proposed deep architecture that implements
the complete MRAN model by realizing the region-level feature extractors
$\phi_m, m = 1, \ldots, M$, the per-source conditional attention estimators
$\omega_m, m = 2, \ldots, M$, and the classifier $C$ by using deep neural
networks, and explain how we jointly learn these networks in an end-to-end fashion.

\begin{figure*}[t!]
\centering
\includegraphics[width=\linewidth]{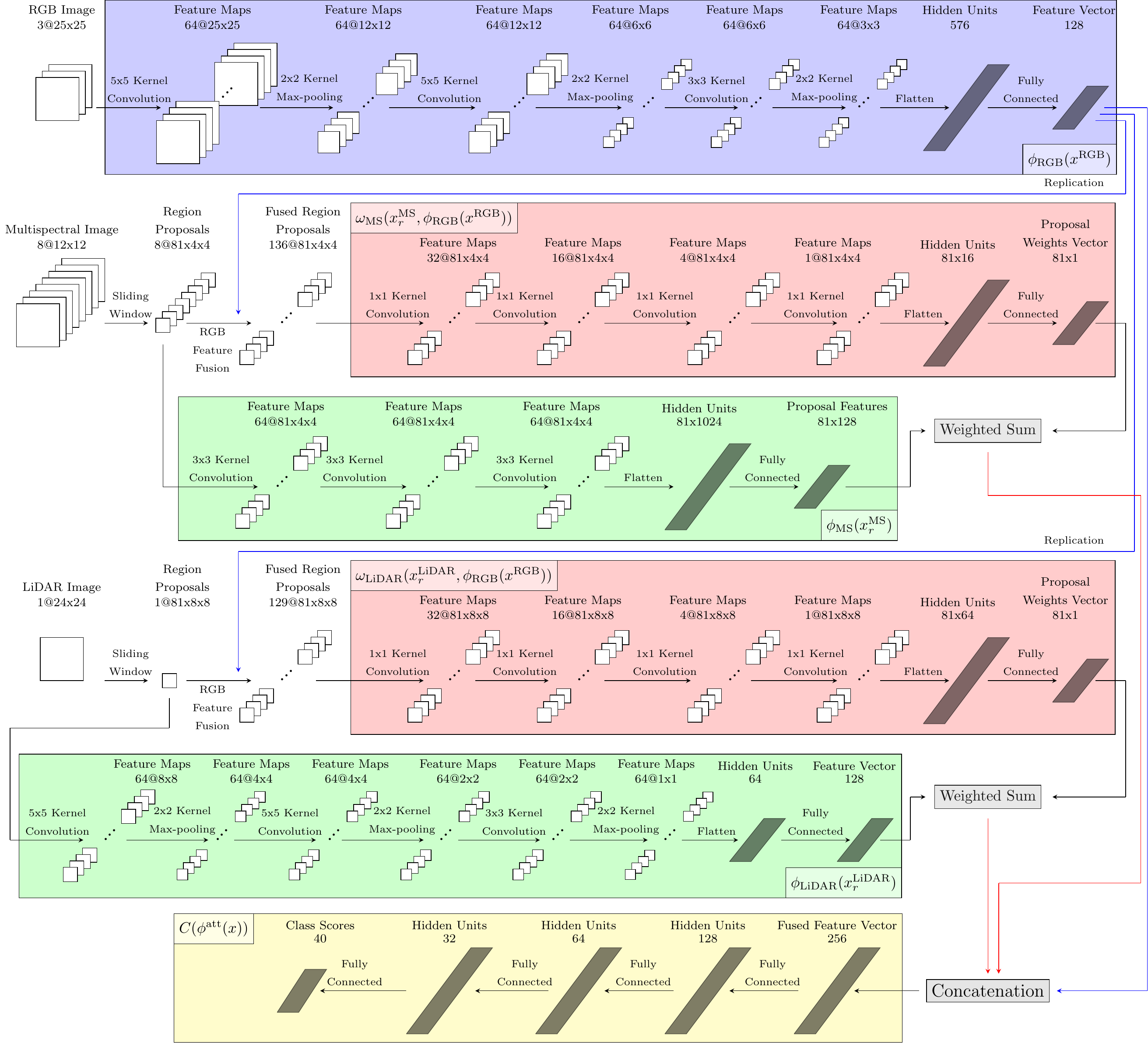}
\caption{Realization of the proposed MRAN architecture by using three source
domains. The colors of the branches correspond to the boxes in Figure
\ref{fig:weight_model}. The first branch $\phi_\text{RGB}$ acts as the
feature extractor for the reference domain, the aerial RGB image $x^\text{RGB}$.
It contains three convolutional layers containing $64$ filters with sizes
$5 \times 5$, $5 \times 5$, and $3 \times 3$, respectively, followed by a
fully-connected (FC) layer containing $128$ neurons. We apply max-pooling after each
convolutional layer. The second branch $\omega_\text{MS}$ estimates the attention
scores for the region proposals $x_r^\text{MS}$ of the MS image, with the help
of the feature representation of the RGB image. This branch contains four
convolutional layers with $32$, $16$, $4$, $1$ filters each with size $1 \times 1$,
followed by an FC layer containing $16$ and $1$ neurons, respectively.
The third branch $\phi_\text{MS}$ computes the feature representations of the
MS region proposals, and includes three convolutional layers each containing
$64$ filters with size $3 \times 3$, followed by an FC layer
containing $128$ neurons. Similarly, the fourth branch $\omega_\text{LiDAR}$
estimates the attention scores for the LiDAR data, and the fifth branch
$\phi_\text{LiDAR}$ computes the LiDAR feature representations.
The last branch $C$ calculates the class scores from the concatenation of the
feature representations of all three sources $\phi^\text{att}$. It consists of
four FC layers containing $128$, $64$, $32$ and $40$ neurons, the
last one giving the class scores. Note that, the feature map sizes and
descriptive names are stated at the top of each layer.}
\label{fig:neural_net}
\end{figure*}

\subsection{MRAN architecture details}
\label{sec:neural_net}

We utilize a deep neural network in order to realize our MRAN framework.
Our goal here is to
(i) jointly process spatial and spectral information in the source images,
(ii) implement an effective conditional region attention mechanism, and,
(iii) learn the whole recognition pipeline in an end-to-end fashion. While the
proposed architecture can easily be adapted to various combinations of sources,
we assume that the reference source is RGB imagery, and the additional sources
are obtained using MS and LiDAR sensors in this presentation.

For this purpose, we define an architecture that is formed by the combination
of five deep convolutional neural network branches and a block of fully-connected (FC)
layers as shown in Figure \ref{fig:neural_net}. The first branch extracts
the image feature representation of the reference RGB data. We adopt this
architecture from our previous work \cite{Sumbul:2018}. The second and
fourth branches take the region proposals of the images from the additional
sources, and append the feature vector of the reference source to the end of
each pixels' input channels via replication. Four convolutional layers with
$1 \times 1$ dimensional filters and an FC layer estimates the
attention scores of region proposals. The third branch in which the feature
representation is computed for each region proposal in the MS data differs
from the first branch for the RGB data by using smaller filters and not using
max-pooling because of the difference in spatial resolution. The fifth branch
is the feature extractor for the LiDAR data and is similar to the first branch.
Finally, the concatenation of attention-driven source representations
and the reference source representation goes to the last branch
in which four FC layers implement the classifier that gives the
final class scores. Stride for all convolutional layers is set at $1$ to prevent
information loss. We use zero-padding to avoid reduction in the
spatial dimensions over convolutional layers.

The number of filters for each convolutional layer in the first, third and fifth
branches is selected as $64$ in order to find a balance between model capacity
and preventing overfitting. However, for the attention score estimator branches,
we prefer to use decreasing number of filters from $32$ to $1$ in order to have
correct number of scores at the end. Finally, although we have experimented with
deeper and wider models, we reached the best performance with the presented network.

The particular instantiation of the network in Figure \ref{fig:neural_net} uses
the RGB data as the reference and MS and LiDAR data as the additional sources.
Note that the \emph{region attention estimator} branches (red boxes) are the
same for all sources, and the \emph{feature extractor} branches (blue and green
boxes) differ only slightly in terms of the number of layers according to the
spatial resolutions of the sources and the sizes of the region proposals. The
design in Section \ref{sec:MRAN} and the abstraction in Figure \ref{fig:weight_model}
are generic so that any number of reference and additional sources with any
spatial and spectral resolution can be handled in the proposed framework by
selecting an appropriate feature extractor model for each source.

Training the model is carried out over the classes by employing the cross-entropy
loss, corresponding to the maximization of label log-likelihood in the training
set. For enhancement of training, we benefited from dropout regularization
and batch normalization. Additional training details and a comparison of our
model are provided in \sect{Experiments}.

%% file: experiments.tex
\section{Experiments}
\label{sec:Experiments}

In this section, we present the experimental setup, results when the sources
are used individually and in different combinations, and comparisons with the
baseline approach.

\subsection{Experimental setup}
\label{sec:expsetup}

We follow the same class split in \cite{Sumbul:2018} and evaluate our method
using both $18$ and $40$ classes. For both cases, we split images from all
sources into {\em train} ($60\%$), {\em validation} ($20\%$) and {\em test}
($20\%$) subsets. Based on our previous observations, we add perturbations to
training images by shifting each one randomly with an amount
ranging from zero to $20\%$ of height/width.

For all experiments, training is carried out on the train set by using
stochastic gradient descent with the Adam method \cite{Adam:2014} where the
hyper-parameters are tuned on the validation set. All network parameters are
initialized randomly and are learned in an end-to-end fashion. The initial
learning rate of Adam, mini-batch size, and $\ell_2$-regularization weight are
set to $10^{-3}$, $100$, and $10^{-5}$, respectively, as in \cite{Sumbul:2018}.

We use normalized accuracy as the performance metric where the per-class
accuracy ratios are averaged to avoid biases towards classes with a large
number of examples.

\begin{table}
\centering
\caption{Single-source fine-grained classification results (in \%)}
\label{table:single_source}
\setlength{\tabcolsep}{2pt}
\begin{adjustbox}{width=\linewidth}
\begin{tabular}{lcccccc}
\cmidrule[.8pt]{2-7}
& \parbox{0.1\linewidth}{\centering Random guess}
& \parbox{0.13\linewidth}{\centering LiDAR\\$8\times8$ patches}
& \parbox{0.13\linewidth}{\centering LiDAR\\$24\times24$ patches}
& \parbox{0.13\linewidth}{\centering RGB\\$25\times25$ patches}
& \parbox{0.13\linewidth}{\centering MS\\$4\times4$ patches}
& \parbox{0.13\linewidth}{\centering MS\\$12\times12$ patches}\\
\midrule
\textit{\textbf{18 classes}} & 5.6 & 12.1 & 25.6 & 34.6 & 39.0 & \textbf{47.7}\\
\midrule
\textit{\textbf{40 classes}} & 2.5 & 7.8 & 18.1 & 23.9 & 25.1 & \textbf{34.6}\\
\bottomrule
\end{tabular}
\end{adjustbox}
\end{table}

\subsection{Single-source fine-grained classification}
\label{sec:single_source}

Our initial experiments consist of evaluating the performance of each source
individually. For this, we use the first, third, and fifth branches of the
network in Figure~\ref{fig:neural_net} for RGB, MS, and LiDAR data, respectively.
We add one more fully-connected layer that maps the output of the last layer in
each branch to class scores to obtain three separate CNNs for single-source
classification.
CNN for RGB data always takes $25 \times 25$ pixel patches as input. CNN for MS
data takes both $4 \times 4$ and $12 \times 12$ pixel patches (corresponding to
green and white squares in Figure~\ref{fig:Alignment}, respectively) as input
in two separate experiments. CNN for LiDAR data is also given $8 \times 8$ and
$24 \times 24$ pixel patches. Each CNN operates on the whole single patch as
there is no region proposal in this setup.

Performances of different sources are summarized in Table~\ref{table:single_source}.
Results show that all settings are clearly better than the random guess baseline
(choosing one of the classes randomly with an equal probability).
Similar trends are observed for $18$-class and $40$-class classification,
though the latter proved to be a more difficult problem as expected.
We also see that the height and limited structure information in LiDAR data
cannot cope with the spectral information in the other sources while the
MS data outperform all the others. Even though MS has one sixth of the
spatial resolution of the RGB data, its rich spectral content proves to be the
most informative for fine-grained classification of trees. We also see that
using larger patches results in higher accuracies. Although $4 \times 4$
patches for MS and $8 \times 8$ patches for LiDAR perfectly coincide with the
point-based ground truth locations that were verified with respect to the RGB
data, the samples for which these patches could not include most of the target
trees due to alignment problems have better predictions when the extended
context in larger patches is used. However, using such patches has a risk of
including irrelevant details in the feature representation. We show that
finding the correct patch with the correct size in the surrounding neighborhood
significantly improves the accuracies in the following section.

\subsection{Multisource fine-grained classification}
\label{sec:multi_source}

In this section, we evaluate the proposed framework (Section~\ref{sec:MRAN})
against the basic multisource model that uses simple feature concatenation
(Section~\ref{sec:basic_model}). During the end-to-end training of the proposed
model, first, the network in Figure~\ref{fig:neural_net} accepts images from
all sources and produces class scores, and then, back-propagation is carried
out with respect to the calculated loss from class labels. The goal is to
simultaneously learn both the spatial distribution of object regions
and the mapping from multiple sources to class probabilities.

\begin{figure}
\centering
\includegraphics[width=0.9\linewidth]{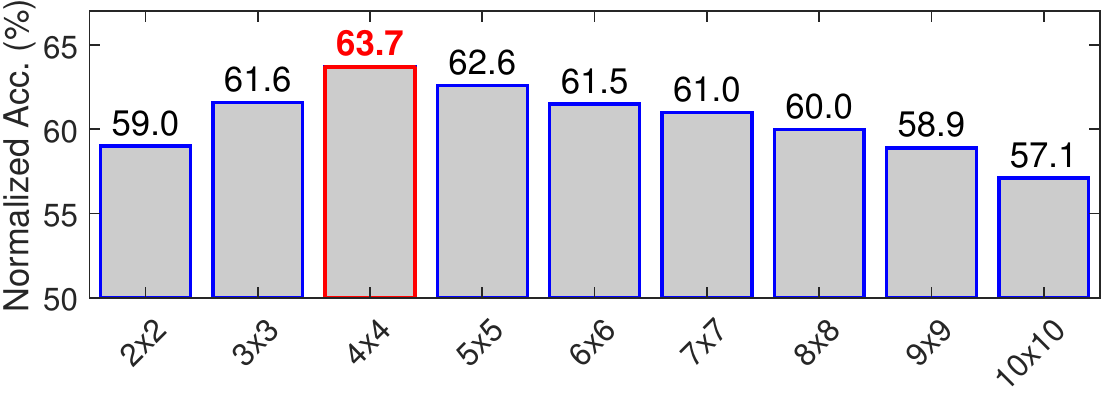}
\caption{Effect of region proposal size on classification performance. The
y-axis shows the normalized accuracy when the RGB data are used with the MS
data as the additional source. The x-axis shows different sized region
proposals within a $12 \times 12$ neighborhood.}
\label{fig:region_proposal}
\end{figure}

\begin{table}
\centering
\caption{Multisource fine-grained classification results (in \%)}
\label{table:multi_source}
\setlength{\tabcolsep}{2pt}
\begin{adjustbox}{width=\linewidth}
\begin{tabular}{lcc}
\toprule
& \textit{\textbf{18 classes}} & \textit{\textbf{40 classes}}\\
\toprule
Random guess & 5.6 & 2.5\\
\midrule
Basic CNN model (RGB \& MS) & 55.5 & 39.1\\
\midrule
Basic CNN model (RGB, MS \& LiDAR) & 56.8 & 41.4\\
\midrule
Recurrent attention model (RGB \& MS) \cite{Fu:2017} & 58.7 & 41.6\\
\midrule
Recurrent attention model (RGB, MS \& LiDAR) \cite{Fu:2017} & 58.2 & 42.6\\
\midrule
Proposed framework (RGB \& MS) & 63.7 & 46.6\\
\midrule
Proposed framework (RGB, MS \& LiDAR) & \textbf{64.2} & \textbf{47.3}\\
\bottomrule
\end{tabular}
\end{adjustbox}
\end{table}

To identify the locations that are likely to contain an object of interest,
our model evaluates sliding windows of region proposals within larger
neighborhoods. We experimented with different region proposal and neighborhood
sizes. Figure~\ref{fig:region_proposal} shows the details of these experiments
when the RGB data and MS data are used together (first, second, third, and sixth
branches in Figure~\ref{fig:neural_net}). Different sized region proposals
were searched within a $12 \times 12$ pixel neighborhood in the MS data, and
$4 \times 4$ pixel regions achieved the best performance. This size is also the
matching spatial dimension of objects when the spatial resolution of the RGB
data is considered. Note that, different sizes lead to different number of
region proposals. For the particular case of $4 \times 4$ regions within
$12 \times 12$ neighborhoods, we obtain $81$ proposals with a stride of $1$
pixel. In the rest of the section, we present the multisource classification
results when $4 \times 4$ regions in $12 \times 12$ neighborhoods are used for
the MS data and $8 \times 8$ regions in $24 \times 24$ neighborhoods are used
(with a stride of $2$ pixels to similarly obtain $81$ proposals) for the LiDAR
data in the proposed framework. Figures~\ref{fig:Alignment} and
\ref{fig:neural_net} also illustrate this particular setting.

Table~\ref{table:multi_source} summarizes the results for multisource
classification for both $18$-class and $40$-class settings. We used two versions
of the basic multisource model in Figure~\ref{fig:basic_model}. The version
named \textit{basic CNN model} uses the first, third, and fifth branches in
Figure~\ref{fig:neural_net} as the feature extractor networks, concatenates
the resulting feature representations, and uses an FC layer as the
classifier. This model is also learned in an end-to-end fashion.
The version named \textit{recurrent attention model} uses a network that learns
discriminative region selection and region-based feature representation at
multiple scales \cite{Fu:2017}. It uses a CNN for feature extraction at each
scale, and an attention proposal network between two scales predicts the
bounding box of the region given as input to the next scale. The network is
trained by a multi-task loss: an intra-scale classification loss that
optimizes the convolution layers, and an inter-scale pairwise ranking loss
that optimizes the proposal network. The final multi-scale representation is
constructed by concatenating the output of a specific FC layer at
each scale. A two-scale network is observed to perform better than a three-scale
one in our experiments. We use the two-scale architecture to train a feature
extractor for each source, concatenate the resulting feature representations,
and train an FC layer as the classifier.
For both versions, the best performing single-source settings
in Table~\ref{table:single_source} ($12 \times 12$ patches for MS and
$24 \times 24$ patches for LiDAR) are used as the inputs to the model. 

As seen in Table~\ref{table:multi_source}, all settings performed significantly
better than the random baseline and the single-source settings. This shows the
significance of using multisource data in the challenging fine-grained
classification problem. When we compare the two additional sources, MS and
LiDAR, the contribution of the MS data in the overall accuracy is more
significant as the rich spectral content appears to be more useful for
discriminating the highly similar fine-grained categories. Overall, we observe
that the proposed framework that simultaneously learns the feature extraction,
attention, and classification networks performs significantly better than the
commonly used basic multisource model with direct feature concatenation.
Considering the observation that using larger patches gives higher accuracies
for single-source classification in Table~\ref{table:single_source}, the
difference between the accuracy of the basic model (e.g., $56.8\%$ for $18$ classes
and $41.4\%$ for $40$ classes for RGB, MS \& LiDAR sources) that uses the larger
patches and the proposed one (e.g., $64.2\%$ for $18$ classes and $47.3\%$ for
$40$ classes) that assigns object localization confidence scores to smaller sized
region proposals via the region attention estimators and uses the resulting
attention-driven multisource feature representations confirms the importance
of learning both the alignment and the classification models.

\begin{figure}
\centering
\includegraphics[width=0.9\linewidth]{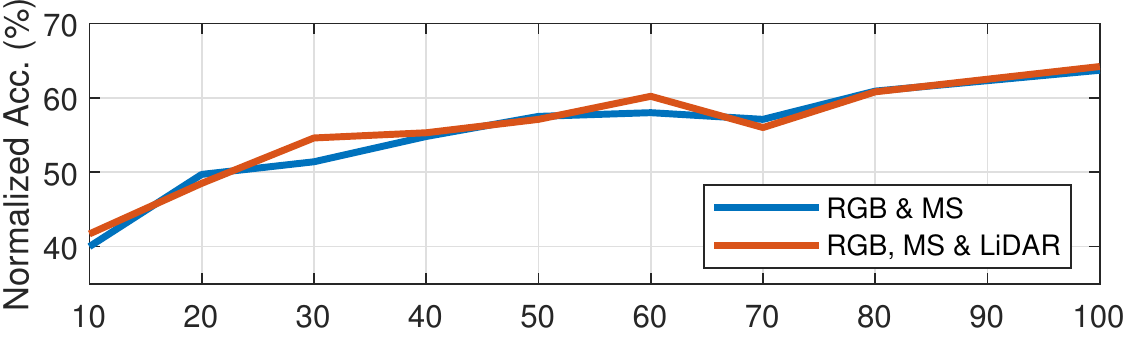}
\caption{Effect of amount of training data on classification performance.
The y-axis shows the normalized accuracy and the x-axis shows the percentage
of the data used during training.}
\label{fig:training_size}
\end{figure}

We also performed controlled experiments to analyze the effect of the amount
of training data on classification performance. Figure \ref{fig:training_size}
shows the resulting accuracies for the $18$-class setting when the amount of
training data is reduced from $100\%$ to $10\%$ with $10\%$ decrements. We
observe that the accuracy achieved by the proposed framework using $50\%$ of
the training data is still higher than that of the basic multisource model using
$100\%$ of the training data. The proposed framework using only $20\%$ of the
training data also performed better than the best single-source model with
$100\%$ of the training data.

When the confusion matrices are considered, we observed that most confusions
are among the trees that belong to the same families in higher levels of the
scientific taxonomy (given in \cite{Sumbul:2018}). For example, among $40$
classes, $30\%$ of thundercloud plum samples are wrongly predicted as cherry
plum, and $16\%$ are wrongly predicted as blireiana purpleleaf plum. Similarly,
$31\%$ of cherry plum samples are wrongly predicted as thundercloud plum.
As other examples for the cases with the highest confusion,
$21\%$ of double Chinese cherry are wrongly predicted as Kwanzan flowering
cherry and $13\%$ are wrongly predicted as autumn flowering cherry,
$21\%$ of common hawthorn are wrongly predicted as English midland hawthorn,
$23\%$ of red maple are wrongly predicted as sunset red maple, and
$18\%$ of scarlet oak are wrongly predicted as red oak.
Since these trees are only distinguished with respect to their sub-species
level in the taxonomy and they have almost the same visual appearance,
differentiating them even from ground-view images with a high accuracy is too
difficult.

\subsection{Fine-grained zero-shot learning}
\label{sec:zsl}

We also evaluate the proposed approach in the zero-shot learning (ZSL) scenario
\cite{Sumbul:2018} where new unseen classes are classified using a model that
is learned by using an independent set of seen classes. Thus, no samples for
the target classes of interest exist in the training data. We follow the same
methodology as in \cite{Sumbul:2018} except the way how image embeddings are
obtained. In place of the feature representation that is obtained from a single
CNN that is trained on RGB data in \cite{Sumbul:2018}, the multisource image
embedding in this paper is obtained from the output of the first fully-connected
layer in the classifier (last) branch of the network in Figure~\ref{fig:neural_net}.
We also evaluate the performances of using feature representations similarly
obtained from the networks trained for the basic multisource model and the
individual single-source models. The class split ($18$ training, $6$ validation,
$16$ test) and the rest of the experimental setup are the same as in \cite{Sumbul:2018}.

\begin{table}
\centering
\caption{Zero-shot learning results (in \%)}
\label{table:zsl}
\setlength{\tabcolsep}{4pt}
\renewcommand{\arraystretch}{1.2}
\begin{tabular}{lc}
\toprule
Image representation & Normalized accuracy\\
\toprule
Random guess & 6.3\\
LiDAR $8\times8$ patches & 8.0\\
LiDAR $24\times24$ patches & 12.1\\
RGB $25\times25$ patches \cite{Sumbul:2018} & 14.3\\
MS $4 \times 4$ patches & 15.2\\
MS $12 \times 12$ patches & 16.7\\
Basic CNN model (RGB \& MS) & 15.8\\
Basic CNN model (RGB, MS \& LiDAR) & 17.4\\
Proposed framework (RGB \& MS) & \textbf{17.7}\\
Proposed framework (RGB, MS \& LiDAR) & 17.0\\
\bottomrule
\end{tabular}
\end{table}

Comparison of different representations evaluated using the $16$ ZSL-test
classes is shown in Table~\ref{table:zsl}. (Additional comparisons with other
ZSL models can be found in \cite{Sumbul:2018}.) When the single-source results
are considered, we observe the same trend as in Table~\ref{table:single_source}
where MS-based representation performed better than LiDAR-based and RGB-based
representations, and using larger patches had higher accuracies than smaller
patches that have potential alignment problems and limited spatial context.
Regarding the multisource results, the best performance of $17.7\%$ was
obtained by the proposed model trained using
RGB and MS data. The more complex models that use all three sources had
slightly lower performances, probably because of the difficulty of learning in
the extremely challenging ZSL scenario with very limited number of training
samples. Overall, together with the supervised classification results in the
previous sections, the ZSL performance presented here highlights the efficacy
of our proposed multisource framework.

\subsection{Qualitative evaluation}
\label{sec:analysis}

The quantitative evaluation results presented above highlight the remarkable
performance by the proposed multisource approach compared to the baseline
feature-level fusion model commonly used in remote sensing multisource image
analysis. Figure~\ref{fig:weight_dist} provides qualitative results to
investigate how well the model solves the alignment problem by learning to
generate meaningful attention scores for the region proposals.
These examples show that our model is capable of estimating
the correct alignment of images obtained from different sources with imprecise
registration while correctly classifying them. 

\setlength{\fboxsep}{0pt}
\newcommand{\example}[1]{%
\includegraphics[width=0.23\linewidth]{samples/#1_rgb}
\begin{minipage}[b]{0.23\linewidth}
\includegraphics[width=0.48\linewidth]{samples/#1_ms}
\includegraphics[width=0.48\linewidth]{samples/#1_lidar}\\[2pt]
\includegraphics[width=0.48\linewidth]{samples/#1_weight_red}
\includegraphics[width=0.48\linewidth]{samples/#1_lidar_weight_red}
\end{minipage}}

\begin{figure}
\centering
\example{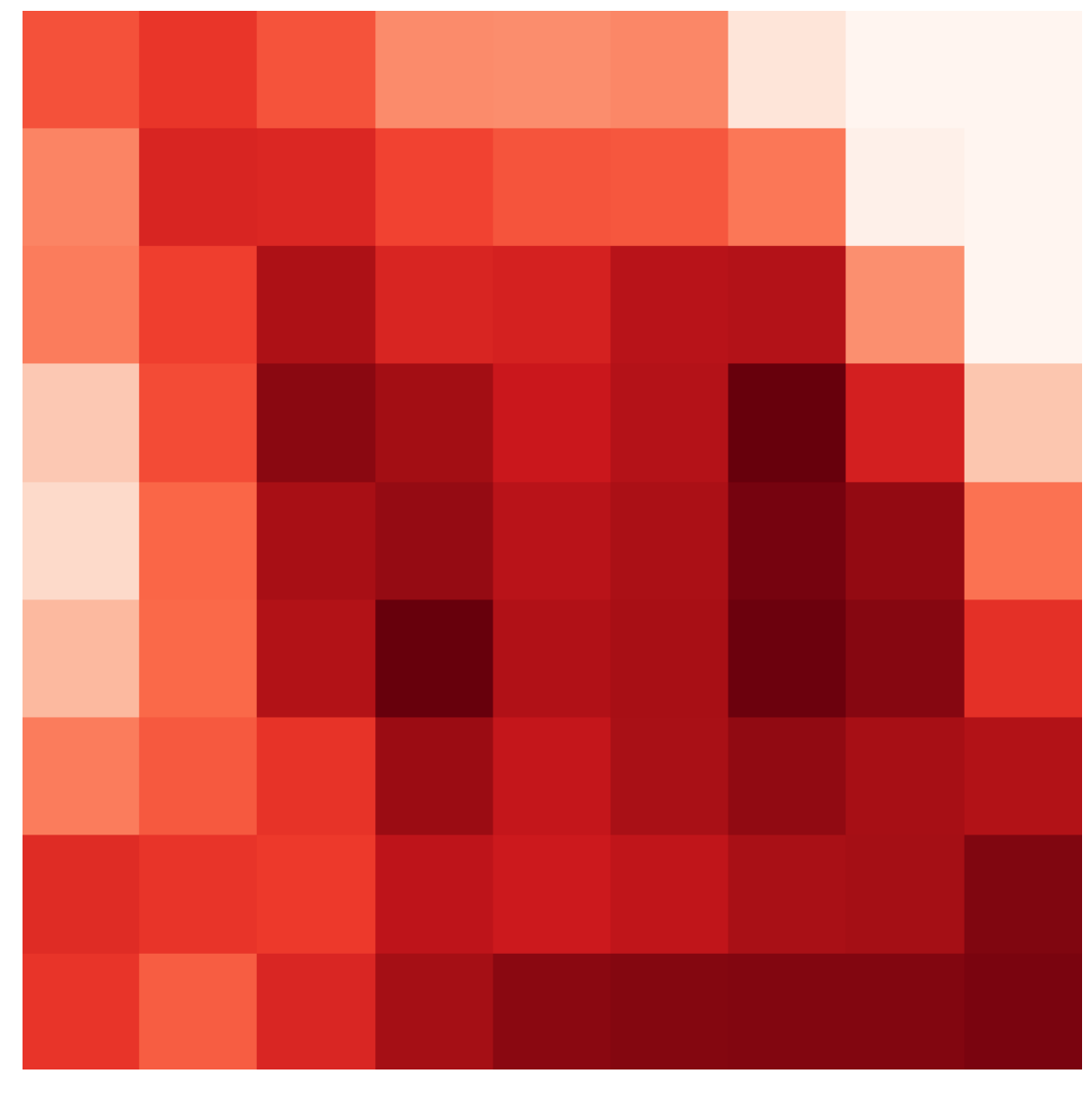} \hspace{1pt}
\example{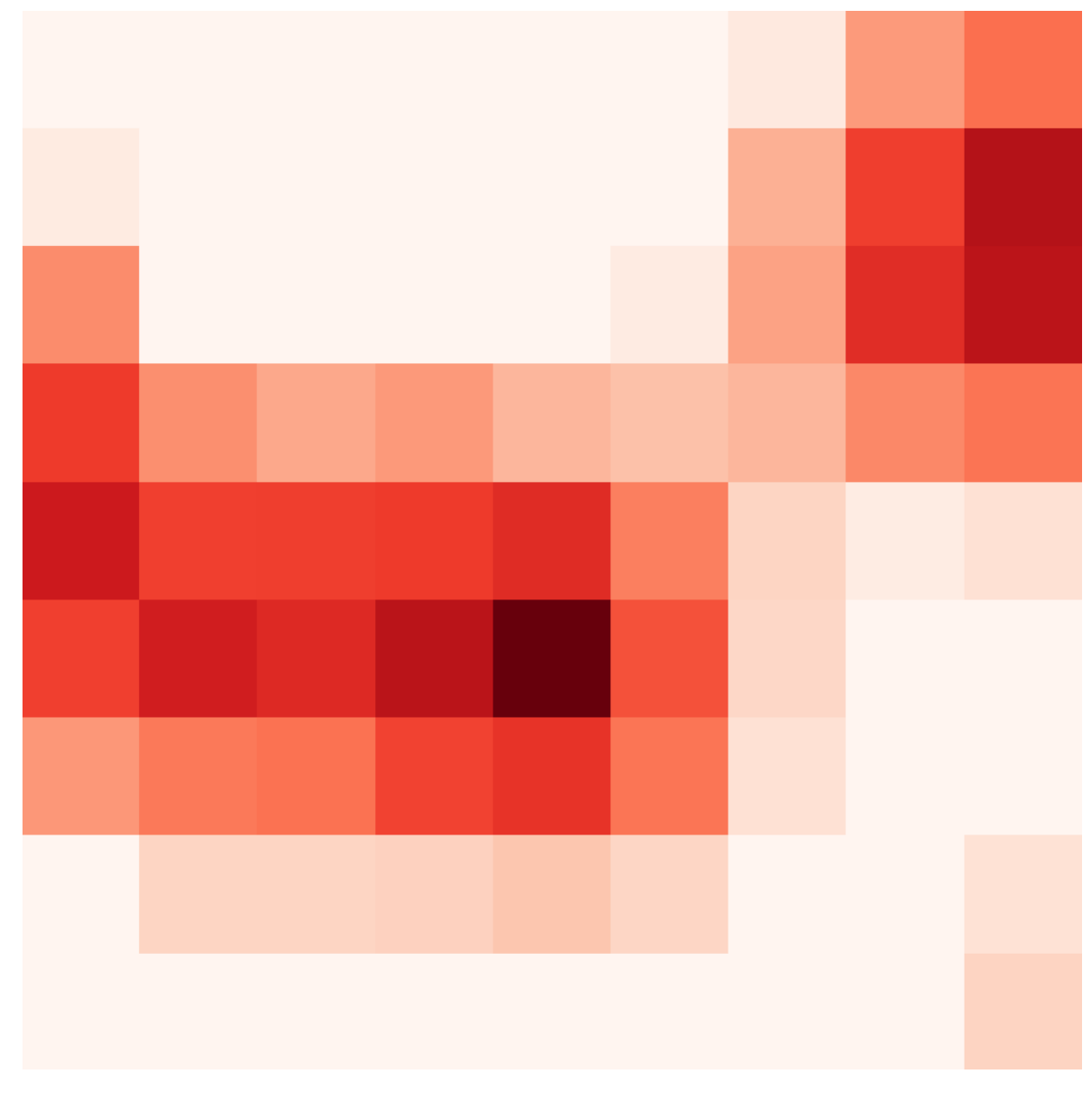}\\[3pt]
\example{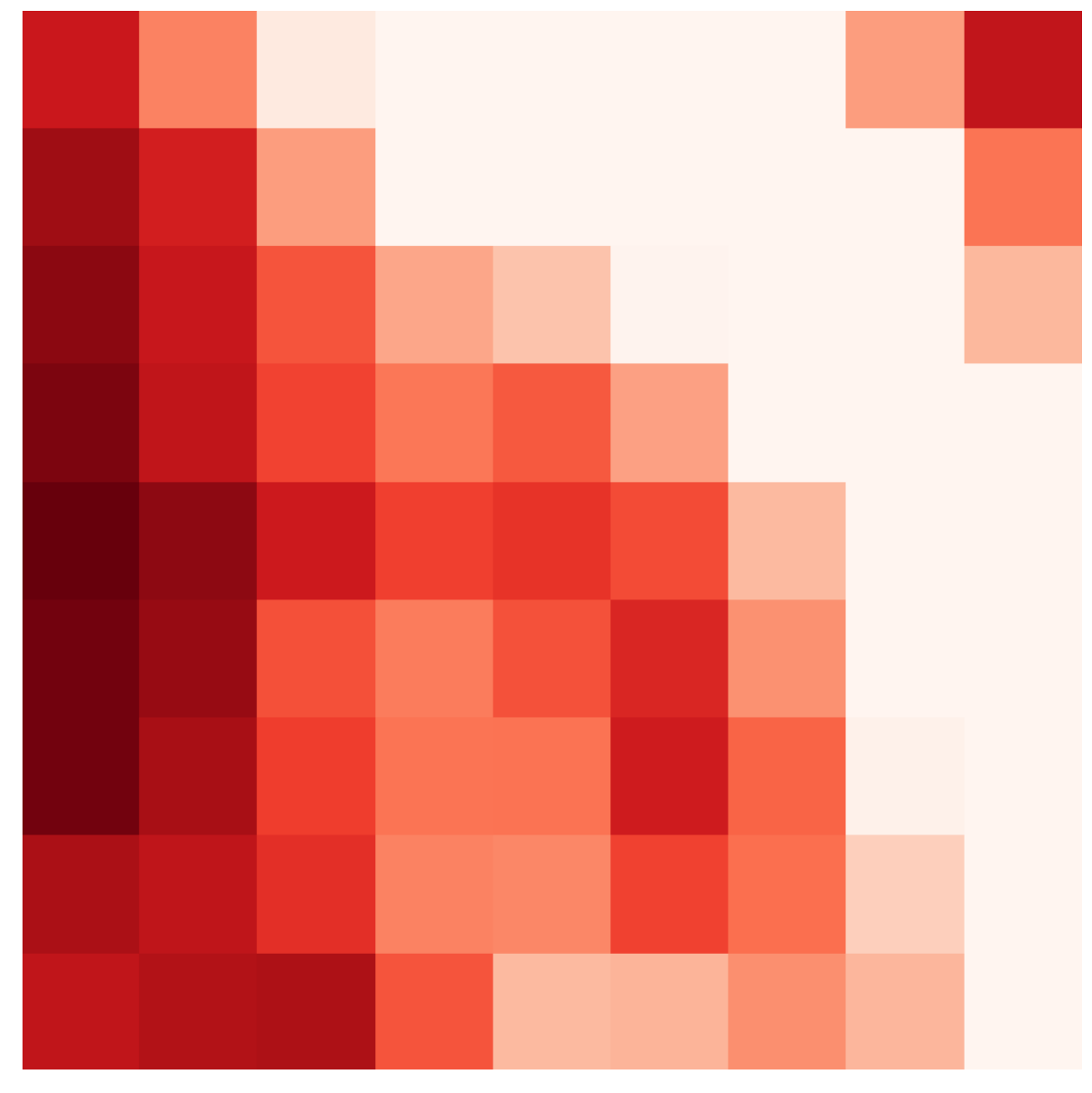} \hspace{1pt}
\example{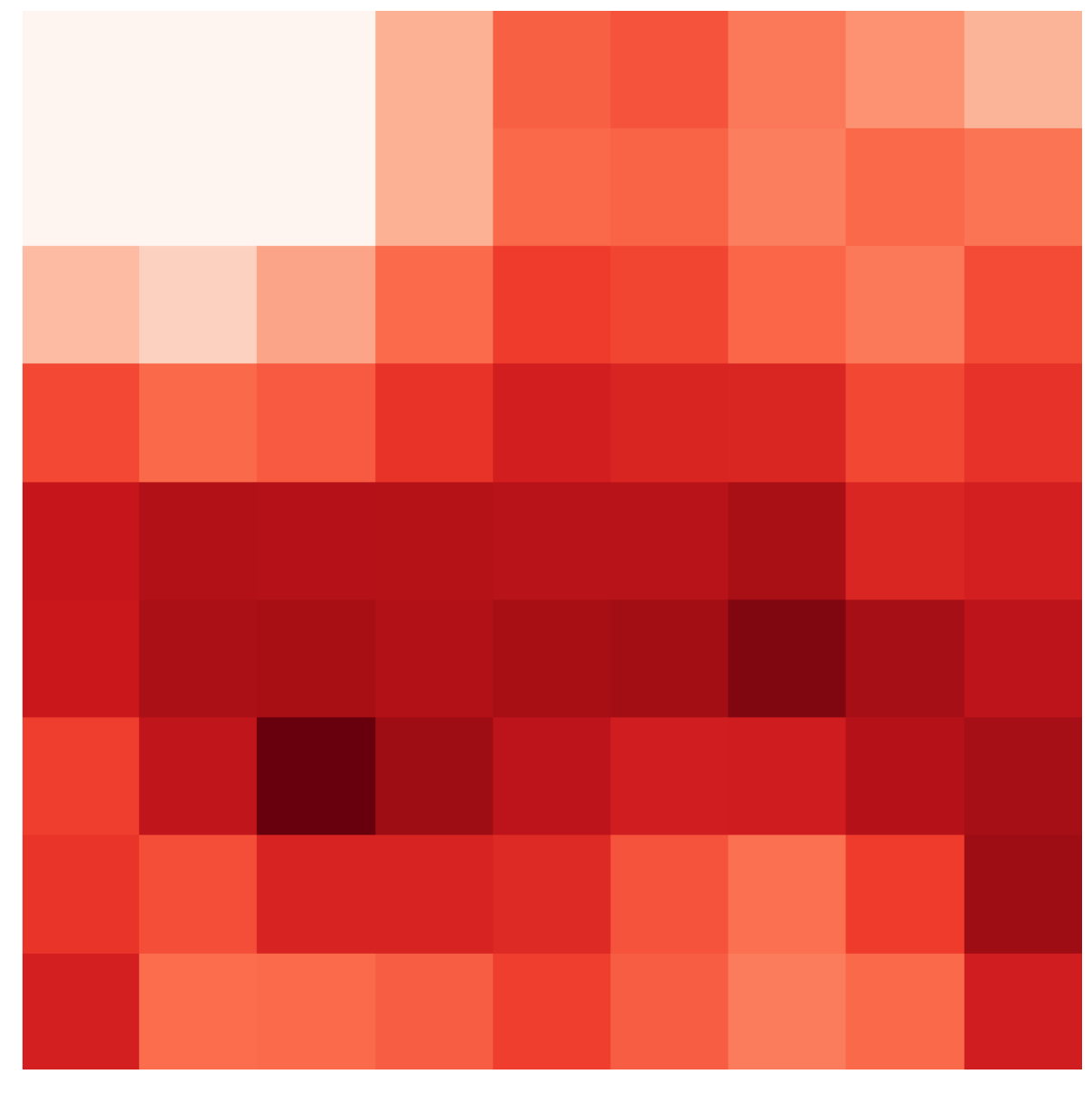}\\[3pt]
\example{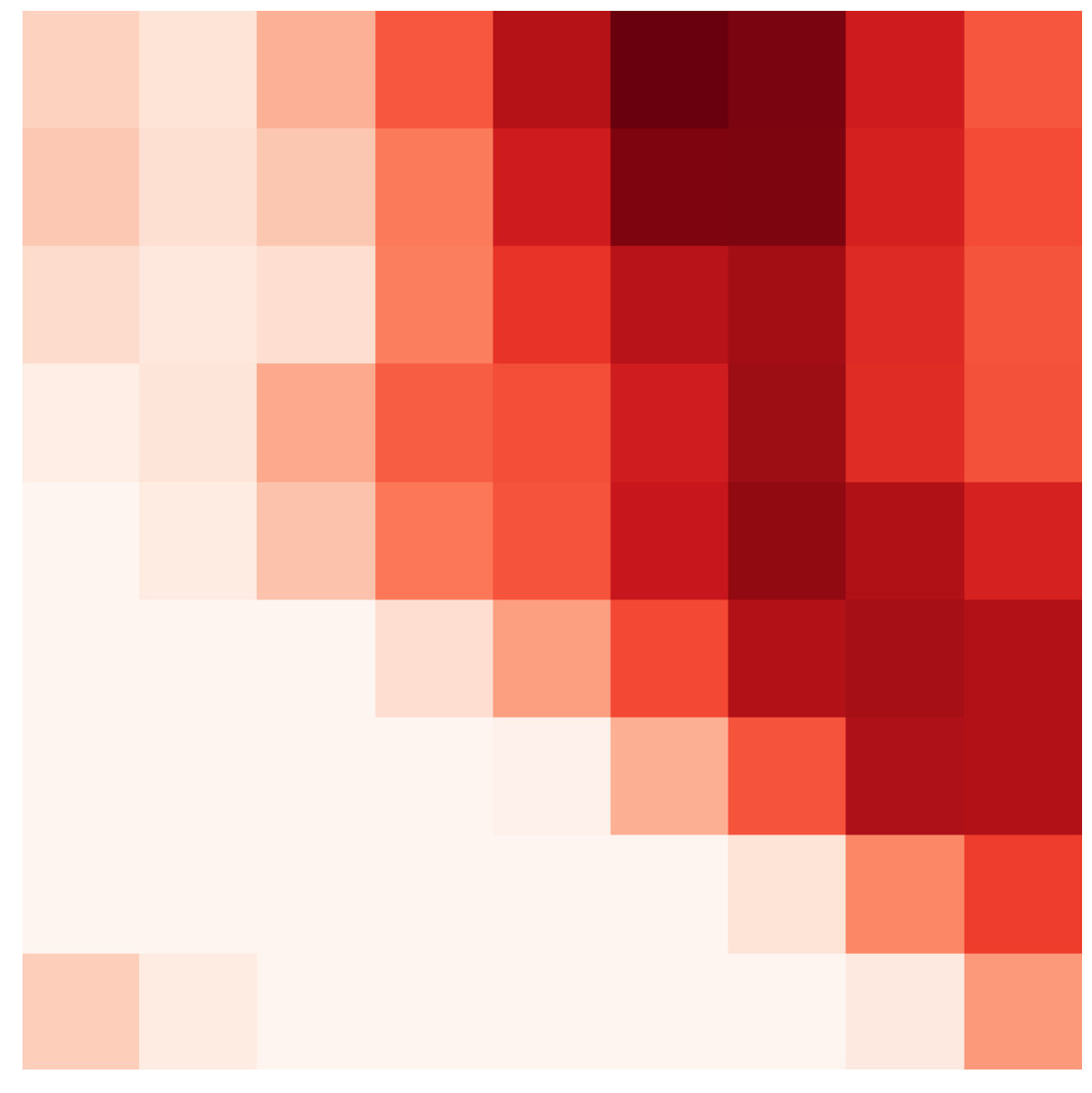} \hspace{1pt}
\example{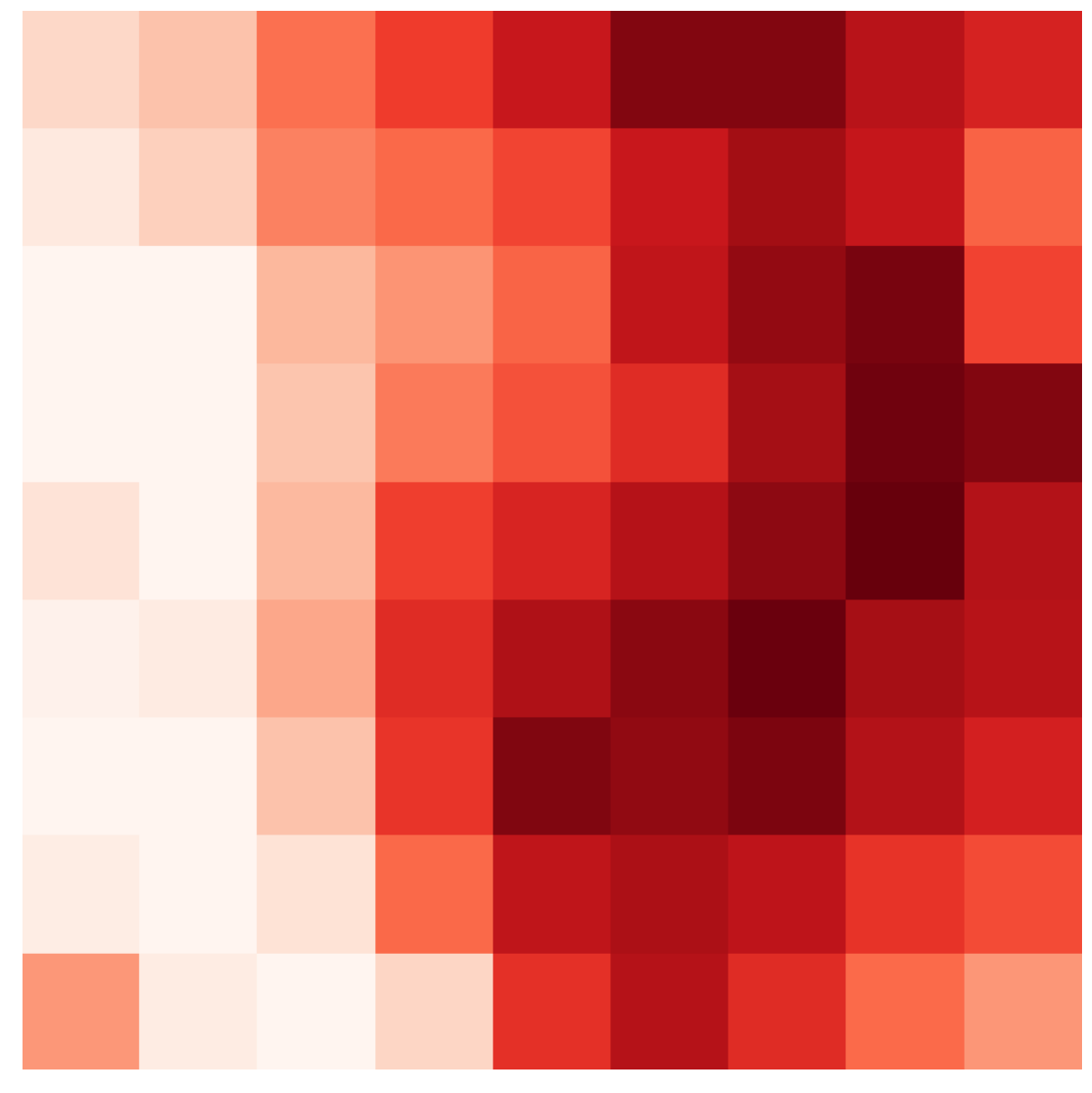}\\
\begin{tikzpicture}[text=black,font=\scriptsize]
\node at (-3cm,-0.2mm) {Colormap:};
\node (map) {\fbox{\includegraphics[width=0.46\linewidth,trim=3 10 3 3,clip]{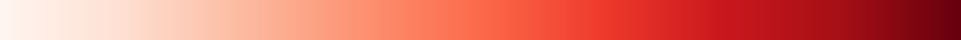}}};
\node at (map.west) {0};
\node at (map.east) {1};
\end{tikzpicture}
\caption{Attention scores for example images. For each sample, the
$25 \times 25$ RGB, $12 \times 12$ MS, and $24 \times 24$ LiDAR patches,
as well as the scores for $4 \times 4$ MS and $8 \times 8$ LiDAR region
proposals within these patches are shown.}
\label{fig:weight_dist}
\end{figure}
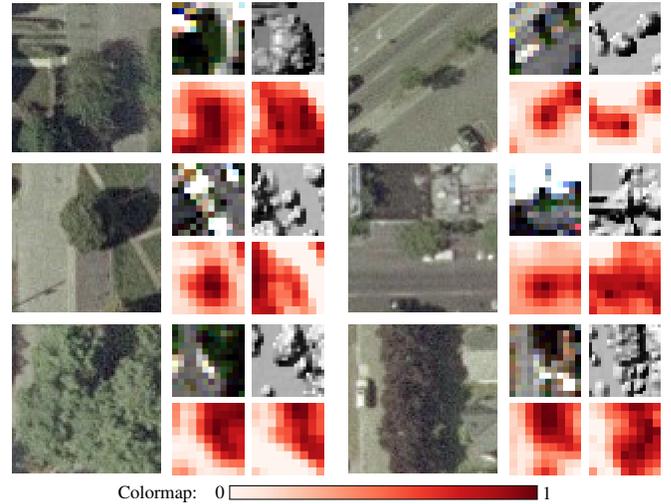

%% file: conclusions.tex
\section{Conclusions}
\label{sec:Conclusions}

We studied the fine-grained object recognition problem in multisource imagery,
potentially having imprecise alignment with each other and with the ground
truth. In order to deal with the complexity of learning many sub-categories
having subtle differences by using multiple image sources with different
spatial and spectral resolutions and with misregistration errors, we proposed
a framework that assigns attention scores to local regions sampled around the
expected location of an object by comparing their content with the features of
the reference source that is assumed to be more reliable with respect to the
ground truth, computes a multisource feature representation as the concatenation
of attention-weighted feature vectors of the local regions, and classifies the
objects using a deep network that learns all of these components in an
end-to-end fashion. Experiments using RGB, MS, and LiDAR data showed that
our approach achieved $64.2\%$ and $47.3\%$ accuracies for the $18$-class and
$40$-class settings, respectively, when all data sources were used, which
correspond to $13\%$ and $14.3\%$ improvement relative to the commonly used
feature concatenation approach from multiple sources. Future work includes
evaluation of the model in other domains, and solving other multisource
classification problems in addition to alignment.

%% file: bio.tex
\vskip -1.5\baselineskip plus -1fil
\begin{IEEEbiography}[{\includegraphics[width=1in,height=1.25in,clip,keepaspectratio]{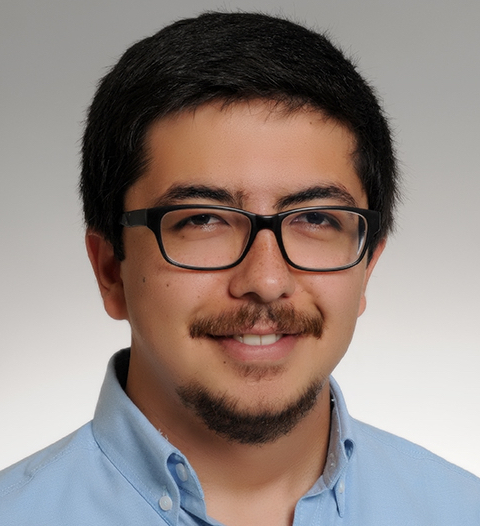}}]{Gencer Sumbul}
	received the B.S. degree in Computer Engineering from Bilkent University,
	Ankara, Turkey, in 2015 and the M.S. degree in Computer Engineering from Bilkent University
	in 2018. He is currently doing a Ph.D. at the Faculty of Electrical Engineering and Computer
    Science, Technical University of Berlin, Germany. His
    research interests include computer vision and machine learning, with special interest in
    deep learning and remote sensing.
\end{IEEEbiography}
\vskip -1\baselineskip plus -1fil
\begin{IEEEbiography}[{\includegraphics[width=1in,height=1.25in,clip,keepaspectratio]{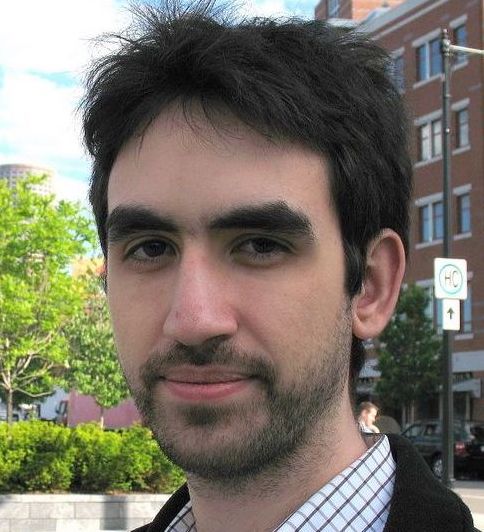}}]{Ramazan Gokberk Cinbis}
    graduated from Bilkent University, Turkey, in 2008, and
    received an M.A. degree from Boston University, USA, in 2010.
    He was
    a doctoral student at INRIA Grenoble,
    France, between 2010-2014, and received a PhD degree
    from Universit\'{e} de Grenoble, France,
    in 2014. He is currently an Assistant Professor at METU, Ankara, Turkey.
    His research interests include machine learning and computer vision,
    with special interest in deep learning with incomplete weak supervision.
\end{IEEEbiography}
\vskip -1\baselineskip plus -1fil
\begin{IEEEbiography}[{\includegraphics[width=1in,height=1.25in,clip,keepaspectratio]{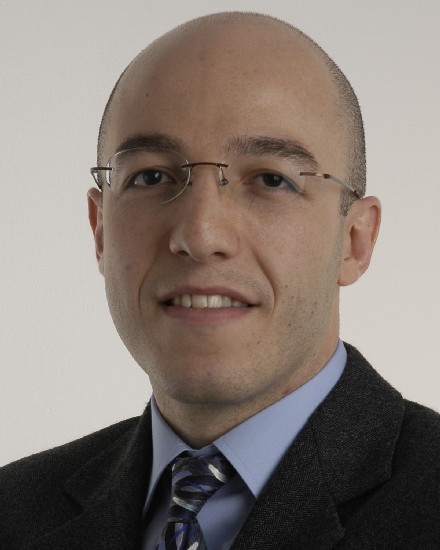}}]{Selim Aksoy}
(S'96-M'01-SM'11) received the B.S.\ degree from the Middle East Technical University,
Ankara, Turkey, in 1996 and the M.S.\ and Ph.D.\ degrees from the University
of Washington, Seattle, in 1998 and 2001, respectively.
He has been working at the Department of Computer Engineering, Bilkent
University, Ankara, since 2004, where he is currently an Associate Professor.
His research interests include computer vision, statistical and structural pattern
recognition, and machine learning with applications to remote
sensing and medical imaging.

\end{IEEEbiography}